# 3D Bounding Box Estimation for Autonomous Vehicles by Cascaded Geometric Constraints and Depurated 2D Detections Using 3D Results

Jiaojiao Fang, Lingtao Zhou, Guizhong Liu
School of Electronic and Information Engineering Xi'an Jiaotong University, China
995541569@qq.com, zhoulingtao7458@hotmail.com, liugz@xjtu.edu.cn

*Abstract*—3D object detection is one of the most important tasks in 3D vision perceptual system of autonomous vehicles. In this paper, we propose a novel two stage 3D object detection method aimed at get the optimal solution of object location in 3D space based on regressing two additional 3D object properties by a deep convolutional neural network and combined with cascaded geometric constraints between the 2D and 3D boxes. First, we modify the existing 3D properties regressing network by adding two additional components, viewpoints classification and the center projection of the 3D bounding box's bottom face. Second, we use the predicted center projection combined with similar triangle constraint to acquire an initial 3D bounding box by a closed-form solution. Then, the location predicted by previous step is used as the initial value of the over-determined equations constructed by 2D and 3D boxes fitting constraint with the configuration determined with the classified viewpoint. Finally, we use the recovered physical world information by the 3D detections to filter out the false detection and false alarm in 2D detections. We compare our method with the state-of-the-arts on the KITTI dataset show that although conceptually simple, our method outperforms more complex and computational expensive methods not only by improving the overall precision of 3D detections, but also increasing the orientation estimation precision. Furthermore our method can deal with the truncated objects to some extent and remove the false alarm and false detections in both 2D and 3D detections.

## I. INTRODUCTION

3D object detection is the task that aims at recovering the 6 Degree of Freedom (DoF) pose and dimensions of interest objects in physical world which are defined by 3D bounding boxes. It is one of the most important task in 3D scene understanding and provides indispensable information for intelligent agents perceptual like autonomous driving vehicles to perceive and interact with the real world. The existing methods are mainly classified into three categories according to the data has been used which falls into LiDAR, stereo image, and monocular image. Although monocular image based methods of most disadvantages as the lack of depth information when projected a 3D scene on image plane, they still have received extensive attention due to its low cost device requirements and wide range of application scenarios.

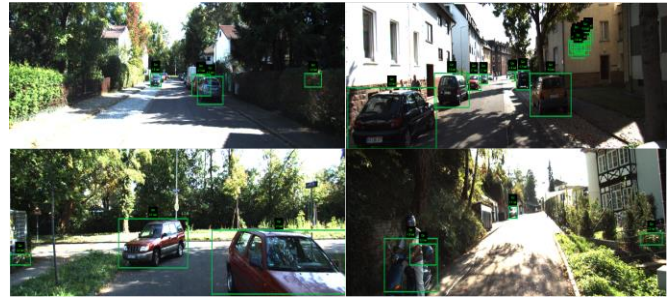

Fig. 1. Examples of some fallacious detection results for car category in the KITTI dataset by the 2D detector used in our paper. From these results we can see that there are many false alarm the 2D detections especially for the upper left image. These wrong detections are harmful for the two stage monocular based 3D object detection methods as their performance are strongly depend on the 2D image understanding.

While deep learning based 2D object detection algorithms [2, 3, 4, 16] have achieved great advance and gain better robustness for challenges such as occlusion, viewpoint variance, illumination variance et.al, it remains an under-constrained problem for 3D object detection based on purely monocular images by back-projecting the object on image plane to the 3D physical world. But this is eventually enabled via deep learning [9] and some geometrical relation between 2D and 3D space [5, 8]. By the data-driven CNN models, we can learn the empirical relationship between objects' appearance and their 3D properties with some specific priors in the scenario. In deep learning based 3D object detection problem, the dimensions and orientation estimation are relatively simple as they are strongly related with object's appearance and can be easily estimated. But it is impractical to directly regress location only by a single image patch, since the depth information is hard to recover only depended on appearance information. The existing location estimation methods try to predict location of the object by the geometric constraint between the 2D and 3D bounding boxes and convert the location estimation to an appearance unrelated problem. However, this conversion has two significant drawbacks: 1) it cannot benefit from a large number of labeled data in the training set; 2) the geometric constraint is hard to select the best fitting configuration and get the optimal solution by solving an over determined equation. Our previous work converts the 2D-3D boxes fitting configuration selecting problem to an appearance related viewpoint classification problem for getting more accuracy configuration selection. All the existing methods are ignored to further utilize the recovered physical world information by 3D detections to further understand the 2D image.

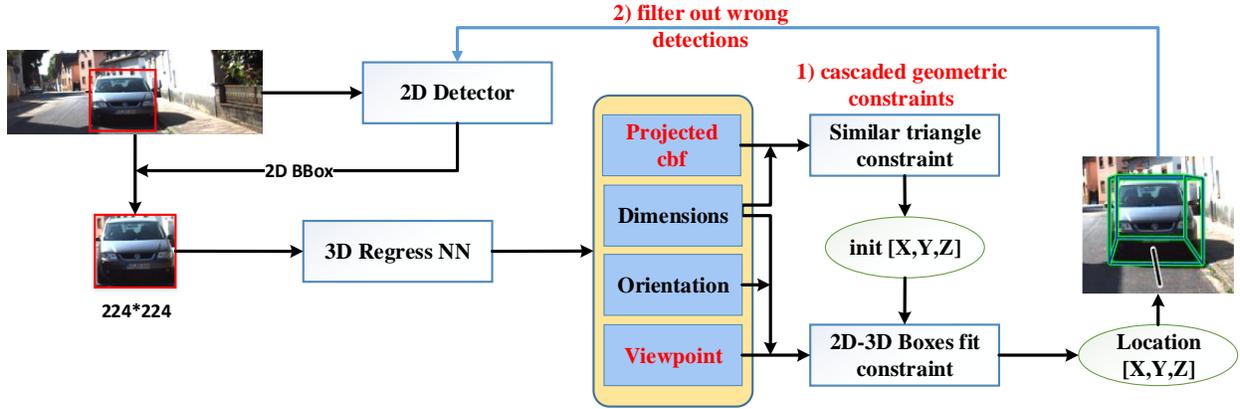

Fig. 2. The overall pipeline of our proposed monocular 3D object detection method, which only requires a single RGB image as input, and can achieve 3D perception of the objects in the scene. Our method use the deep CNN to regress another two appearance-related 3D properties for 3D location. The red fonts represent the most important two key modules, cascaded geometric constraints and 2D detections filtration based on the 3D results.

In this paper, we introduce a novel deep learning and monocular based two stage 3D object detection method by cascaded geometric constraints for location estimation. Firstly, in contrast to most current networks that only regressing the dimensions and orientation of 3D objects, we modify the existing multi-branch deep CNN by adding two branches of viewpoints classification and the center location projection regression of the 3D bounding box's bottom face (CBF) to obtain another two stable appearance related object properties for 3D location estimation. Secondly, we use the estimated CBF combined with the similar triangle constraint between the heights of 2D and 3D boxes to solve an initial 3D location by a closed-form solution. If the objects are truncated by the image plane, this solution is used as the final 3D location. Otherwise we will solve the final 3D location by cascaded another constraint. Thirdly, if the object is not truncated, the solved location will be used as the initial value of an over-determined equation constructed by the 2D-3D boxes fitting constraint with the configuration obtained by the viewpoint of the network output.

Most of the prior knowledge about the physical world is missing when projected a scene on image plane, so it is hard to identify the false alarm and false detection in the 2D detectors without 3D spatial information. However, with the help of the 3D detection results, we can recover the 3D spatial prior information to a certain extent which can help us to distinguish the obviously unreasonable 2D detection results both for false detection and false alarm. Thus we can improve the reliability of the 2D and 3D detections at the same time. Fig. 1 gives an illustration of the false alarm and false detection in our 2D detections.

The main contributions of this paper are three aspects as follows:

1) We propose a novel purely monocular image based 3D object detection method by adding two appearance-related 3D properties to the existing multi-branch CNN for more stable 3D properties and building cascaded geometry constraints. Thus we can simultaneously estimate dimensions, orientation, CBF and viewpoints classification. By jointly training several most related tasks, we can improve their performance at the same time.

2) We use the regressed CBF with the similar triangle constraint to solve the initial location of the 3D objects which is used as the initial values to solve the over-determined equations obtained by the 2D-3D bounding box fitting constraints with the estimated viewpoint to get more accuracy 3D bounding boxes. And our method can deal well with some truncated objects by only use the initial value as the 3D location.

3) We use the prior information of the physical world which is recovered by the 3D detections combined with the image acquisition device, the acquisition environment to distinguish whether the detection is false or not. Thus we can get more reliable 2D and 3D detections.

Experiments on KITTI dataset demonstrate the improvement of our work on orientation estimation and overall detection precision compared with current state-of-the-art methods only using a single RGB image. The qualitative results show that our method can deal well with some truncated objects and filter out the wrong detections in 2D and 3D results.

## II. RELATED WORK

Recently, deep convolutional neural networks (CNN) based 2D object detections have achieved remarkable performance improvement. Thus the more meaningful task 3D object detection which is estimating the 3D bounding boxes of the objects has drawn more and more attention. And plentiful of methods have been proposed to address the problem of 3D object detection by the data collected from autonomous driving scenarios, such as KITTI dataset. The existing 3D object detection methods are mainly classified into three categories as follow.

*A. LiDAR-based 3D Object Detection*

Another kind of 3D detection method uses LiDAR data as the additional information to get more accurate 3D detection result. [11] proposes a method that uses LiDAR data to create multi front view and bird eye view (BEV) of the scene and feed them separately into a two-stage detection network along with the RGB image frame. After proposals are obtained from each view, a multi-stage feature fusion network is deployed to get the 3D properties and classes of objects. In [12], 3D anchor grid is used to get object candidates. After the extracted CNN features of each candidate from both images and LiDAR-based BEV map are fused and scored, top scored proposals are then classified and dimension-refined by the second stage of the network. This kind of method produces much better 3D AP score due to the known depth information, but the cost of money and computing resources as well as the power consumption of devices capturing depth information is not bearable for some circumstances. And these methods are like the way human perceiving the 3D physical world, in which no particular depth information is acquired, either.

*B. Stereo Images based 3D Object Detection*

Stereo vision is also used in some 3D detection algorithms for its simulation to human binocular vision. Based on Mono3D proposed by the same group, in [13], stereo images are used to get better 3D proposal in physical world. Stereo-based HHA feature [14] which encoding the depth information of the scene is also used as a stream of input to get better 3D bounding box regression. [15] proposes a stereo-extended faster R-CNN detection method in which region proposal is done on both left and right image from a stereo pair through RPN and their results are associated. After the keypoints, viewpoints and object dimensions are estimated from stereo proposals, a refinement procedure is deployed via region-based photometric alignment to get better detection.

*C. Monocular Image based 3D Object Detection*

Taking advantage of the prior information from the dataset, in [5], 3D sliding window and a sophisticated proposal method considering context on image, shape, location, and segmentation feature, is employed to generate 3D proposals from scenes. A Fast R-CNN-based detector is fed with image patches correspond to each 3D candidates to classify them and regress the bounding boxes and orientations of objects. [6, 7] separately detect objects that fall in different sub-categories in physical world. The sub-categories are defined by objects' shape, viewpoint and occlusion patterns and divided by clustering using 3D CAD object models. [10] proposes a method called deep MANTA which deploys a Faster-RCNN style model to detect objects and their parts on images from coarse-to-fine. 3D CAD models are also used to establish a library of template to be matched with model's output so that the orientations and 3D positions of objects can be inferred. [8] proposes a 2D-3D object detection framework that regresses objects' orientation and dimensions from image patches containing objects utilizing 2D bounding box produced by efficient 2D detector. 2D-3D box geometric constraints are then be found to calculate the 3D positions of objects. Although the procedure of finding eligible constraints can be done in parallel, it still leads to unnecessary computational and time consumption which limits the application of this method. [18] and [19] use geometric constraint to get an initial 3D bounding box and refine them by a complex post-processing step.

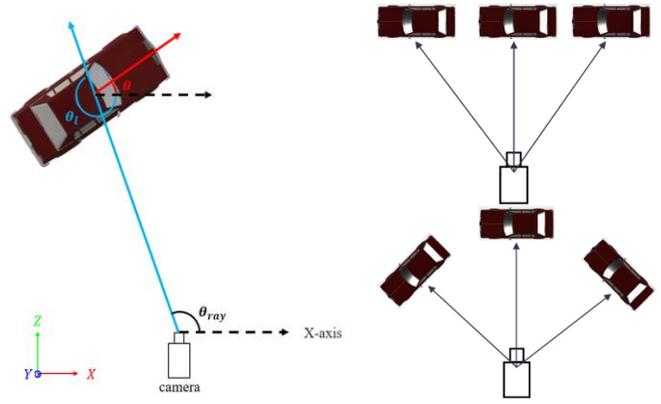

Fig. 3. In the left is an illustration of the global orientation $\theta$, angle of ray $\theta_{ray}$ that from camera to a car's center and local orientation $\theta_l$ which are predict by the network. The upper right shows an example of the cars with the same global angle, but different local angle and the lower right shows the cars with same local angle, viewpoint but different global angle.

## III. 3D Bounding Box Estimation Using Deep Learning and Geometry Constraint

In this paper, we propose a novel two-stage 3D object detection method based on the advanced 2D detector, reliable 3D object properties and geometric constraints between 2D and 3D boxes to recover a complete 3D bounding boxes. In contrast to the existing methods that only regressing orientation and dimensions of 3D object, our method uses a multi-branches deep CNN to regress two additional components which are used to solve the location of the 3D objects by cascaded geometric constraints. At the 2D detection stage, an advanced 2D detector is applied to determine the sizes and positions of objects on image plane. And then the cropped image patches according to 2D detections are fed into a multi-branches CNN to respectively infer: 1) dimensions 2) orientation, i.e. the local angle is regressed to calculate the global angle that only the yaw angle be specified in autonomous driving scenarios in KITTI dataset 3) viewpoint 4) the center projection of the bottom face. We use the estimated dimensions and orientation as well as other two properties combined with the constraints between the 2D-3D boxes to build two sets of linear equations which are used to recover location of the object. Thus we can solve a more precision 3D location by cascaded geometric constraints. Once the 3D bounding boxes are obtained, we use some physical properties to distinguish whether they're reasonable result or not. The overall pipeline of our methods is shown in Fig. 2.

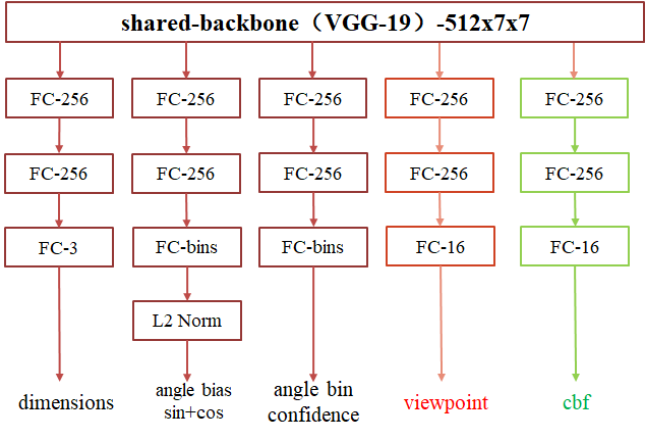

Fig. 4. Architecture of our multi-task network which consists of five branches to compute dimension residual, angle residual, confidence of each bin, viewpoint classification and the center projection of bottom face respectively.

*A. Regressing Two Additional Appearance-related 3D Properties for 3D Location Estimation*

In most of the existing two stage 3D object detection methods, only the dimensions and orientation of the 3D objects are regressed, but the dimensions and orientation are not enough for determining the location of a 3D box. There are several different 3D properties that tied strongly to the visual appearance can be estimated to further constrain the final 3D box estimation.

Inspired by the position estimation in the CNN based 2D detections supervised by the labelled ground truth, we add one task called the center projection position regression to the current dimensions and orientation regression network. The ground truth position of the projection is computed by projecting all the 3D center location of the objects to the image plane with camera intrinsic matrices which will be used for supervision. This prediction can be used in the classical pinhole camera model which provides similar triangle constraint between the sizes of 2D-3D boxes. Given the 2D projection $[uc, vc]$ of a 3D box's location $[X, Y, Z]$, and the 2D box $[u_{min}, u_{max}, v_{min}, v_{max}]$, we use the network to regress the offset between $[uc, vc]$ and $[(u_{min} + u_{max})/2, v_{max}]$ for more reliable results.

Another task called viewpoints classification is also added into this deep CNN. The 3D location estimation method in [8] is not only concept simple but also effective by using the geometric constraint between the 2D-3D bounding boxes fitting. Our previous work using the viewpoint classification task to further improving its performance by roughly dividing 64 kinds of correspondence configurations into 16 categories according to which surfaces of the 3D objects are projected to image plane. We also adopt this method to determine the configuration by classifying viewpoints from where the object is observed. By adding these two tasks, we can transfer the location estimation to an appearance related problem. Thus the deep CNN we used has five sub-branches in total and is trained jointly among all these tasks. The complete architecture of our 3D properties estimation network is shown in Fig. 4.

There are two orientations in the 3D object detection problem, global orientation and local orientation. The global orientation of an object is defined in the world coordinates and will not change with the camera pose, while the local orientation is defined under the camera coordinates and hinges on how the camera shots the object.

Due to the fact that object with same global orientation can look differently after projected on image plane if their spatial position varies, we also regress the local orientation $\theta_l$ instead of directly regressing global orientation $\theta$ as in [8]. As shown in Fig. 3, the regressed local orientation $\theta_l$ is more relevant to the appearance of cropped image patch. And the global orientation can be calculated as:

$$\theta = -(\theta_{ray} + \theta_l - 2\pi) = 2\pi - \theta_{ray} - \theta_l \quad (1)$$

Where $\theta_{ray} = \arctan(z/x) = \arctan(f/(c_v - vc))$ denotes the rotations between the camera principle axis $[0, 0, 1]^T$ and the ray passing through the center of the 3D object which can be computed easily using the camera intrinsic parameters and object's position on the image. $c_v$ is the vertical direction coordinate of the camera principle point.

A MultiBin method [8] that decomposing the continuous orientation angle into discrete bins and regressing residual of each bin is used in our paper. Dimensions' offsets, (Δdx, Δdy, Δdz), which are residual of object's dimensions to their category's average dimensions are predicted for the diversity of dimensions distribution of different class objects. Using discrete bins and predicting offsets from mean sizes facilitates orientation and dimension learning by restricting values to be within a smaller range. To avoid regressing a periodic value of angle $\theta_l$, a L2-norm layer is added at the end of its branch to generate the sine and cosine prediction of residual angle and its loss is defined by the cosine similarity between the prediction and the real residual angle:

$$L_{ang} = 1 - 1/n_\theta \Sigma \cos(\theta^* - c_i - \Delta\theta_i) \quad (2)$$

where $\theta^*$ denotes the ground-truth local orientation, $c_i$ denotes the i-th bin region that ground-truth falls in, $\Delta\theta_i$ is the predicted residual angle and $n_\theta$ is the number of overlapped bins covering the ground-truth and the loss function uses the cosine function to ensure that the offset $\Delta\theta_i$ can be well regressed. Residual regression can significantly reduce the range of a continuous variable and improve the estimation precision.

Thus the overall loss function $L$ can be denoted as:

$$L = w_1 L_{dims} + w_2 L_{ang} + w_3 L_{conf} + w_4 L_{view} + w_5 L_{cbf} \quad (3)$$

where $L_{dims}$, $L_{ang}$, $L_{conf}$ and $L_{view}$ denote the losses for dimension regression, angle bias regression, confidence of bins, viewpoints classification and the center projection of the 3D box's bottom face respectively. $w_{1:5}$ denote the combination weighting factors of each loss.

*B. 2D-3D Boxes Fitting Constraints with Viewpoints Classification*

The fundamental idea of computing 3D location of object by the 2D-3D boxes fitting constraint comes from the consistency of 2D and 3D bounding boxes. Specifically, the projected vertexes of an object's 3D box should fit tightly into each side of its 2D detection box. In other words, four of eight vertexes of the 3D box should be projected right on the four sides of 2D box respectively. In the left of Fig. 5 shows an example of the cars' 2D-3D boxes fitting from KITTI dataset, which shows one kind of correspondence configuration between 2D and 3D box. The vertex numbers 6, 1, 5, 3 of 3D box are projected on the upper, lower, left, right sides of the 2D box respectively, and the viewpoint are from front-right. In total, the sides of the 2D bounding box provide four constraint equations on the 3D bounding box. Given the camera intrinsic matrix $K$, the dimensions of object $d = [l, h, w, 1]^T$, the 2D box $[u_{min}, u_{max}, v_{min}, v_{max}]$ and the global orientation $\theta$, these corresponding boxes fitting configuration constraints can be formulated as:

$$\begin{cases} u_{min} = \pi_u(K[R_\theta\ T]S_1 d) \\ u_{max} = \pi_u(K[R_\theta\ T]S_2 d) \\ v_{min} = \pi_v(K[R_\theta\ T]S_3 d) \\ v_{max} = \pi_v(K[R_\theta\ T]S_4 d) \end{cases} \quad (4)$$

where $R_\theta$ is the rotation matrix parameterized by orientation $\theta$. And $T = [t_x, t_y, t_z]^T = [X, Y, Z]^T$ denotes the transition from the origin of camera coordinates to the center of the 3D bounding box's bottom face which needs to be solved by these equations. $\pi_u$ and $\pi_v$ denote the image coordinates extracting functions getting homogeneous coordinates of the object on the image plane as:

$$\begin{aligned} \pi_u(P) &= p_1/p_3 \\ \pi_v(P) &= p_2/p_3 \\ P &= [p_1, p_2, p_3]^T. \end{aligned} \quad (5)$$

And from $S_1$ to $S_4$ are the vertexes selecting matrixes describing the coordinate positions of four selected vertexes in the original coordinate with respect to the object object's dimensions. These matrixes varied with different constraint configurations are used to define the relationship between 2D and 3D bounding boxes. For the car presented in Fig. 5, its corners selecting matrixes of the 2D box's left, right, top and down edges are as follows:

$$S_{1,2\atop 3,4} = \begin{bmatrix} -0.5 & 0 & 0 & 0 \\ 0 & -1 & 0 & 0 \\ 0 & 0 & -0.5 & 0 \\ 0 & 0 & 0 & 1 \end{bmatrix}, \begin{bmatrix} -0.5 & 0 & 0 & 0 \\ 0 & 0 & 0 & 0 \\ 0 & 0 & 0.5 & 0 \\ 0 & 0 & 0 & 1 \end{bmatrix}, \\ \begin{bmatrix} 0.5 & 0 & 0 & 0 \\ 0 & -1 & 0 & 0 \\ 0 & 0 & -0.5 & 0 \\ 0 & 0 & 0 & 1 \end{bmatrix}, \begin{bmatrix} 0.5 & 0 & 0 & 0 \\ 0 & 0 & 0 & 0 \\ 0 & 0 & -0.5 & 0 \\ 0 & 0 & 0 & 1 \end{bmatrix}. \quad (6)$$

We setup a classification task by CNN to determine which configuration we should use to calculate objects' location according to the appearance information shown on image. As shown in the right of Fig. 5, there are totally 16 kinds of viewpoints and each viewpoint are corresponded with a set of vertexes selecting matrixes.

The viewpoints classification branch shares the backbone network with other tasks can increase the model's sensitivity of orientation estimation. We believe that by adding relevant task can facilitate the training process. Nevertheless, this constraint will be invalid once the object is truncated by the image plane.

*C. 3D Box Location Estimation by Cascaded Constraints*

We use the estimated dimensions, orientation, viewpoint and the center projection to predict location of the 3D object. Our 3D location estimation approach is based on the discovery that using similar triangle constraint can easily obtain a 3D location of an object but just an roughly approximation to the actual location due to perspective projection and viewpoint, while the 2D-3D boxes fitting constraint is a much more reliable constraint than the prior but it needs to solve an over determined linear equations which are hard to get an optimal solution.

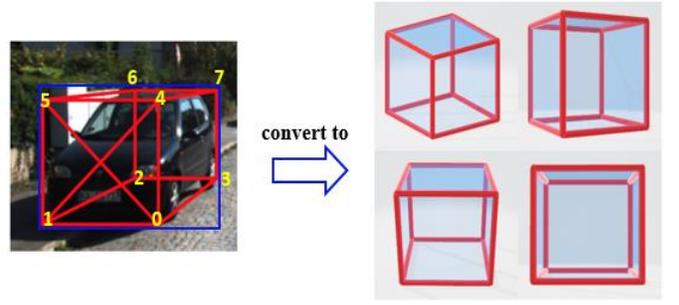

Fig. 5. Illustration of the second geometric constraint we have used. Our previous work convert the left corners selecting problem to the right viewpoint distinguish problem according to the consistency between them. The left part is a sample from KITTI dataset illustrating the relation between 2D and 3D bounding box fitting, where the Arabic numbers indicate the index of the 3D box's vertex. The right part show examples of 4 kinds of different observational viewpoint. From top to bottom are examples of lateral view observation and front view observation. From left to right are examples of look down view and the horizontally view observation.

According to the predicted bottom face's center projection $[uc, vc]$ and the sizes of the 2D box, we can get a linear equations system through equation (7) and (8), of the relationship between 3D object's center points of the bottom face and top face and their projections on image plane.

$$Z \begin{bmatrix} uc \\ vc \\ 1 \end{bmatrix} = K[R\ T] \begin{bmatrix} X \\ Y \\ Z \\ 1 \end{bmatrix} \quad (7)$$

Where $[X, Y, Z]$ represents the 3D location of the object and $[uc, vc]$ is its corresponding projection on image plane. We approximate the center projection of the 3D box's top face as $[uc, v_{min}]$, then

$$Z \begin{bmatrix} uc \\ v_{min} \\ 1 \end{bmatrix} = K[R\ T] \begin{bmatrix} X \\ Y - h \\ Z \\ 1 \end{bmatrix}. \quad (8)$$

Through these, we can get a relation among depth Z from the image plane, focal length f, object height h, and its projection size $vc - v_{min}$ on image plane through similar triangles constraint $(vc - v_{min})/h = f/Z$, as shown in Fig. 6. Thus we can get an initial location of a 3D object, but these are rough solution and rarely equivalent to the real value due to perspective projection, the approximation and camera viewpoint. And this constraint is more sensitive to regression errors, but can be used as reliable initial values to an over-determined equations system for solving a more precision location in the 3D space.

Hence we propose a cascaded geometric constraints method to solve the location of the 3D bounding box. If the objects are truncated by the image plane, the initial location is used as the final solved location. Otherwise we use the Gauss-Newton method to solve these over-determined multivariate equations constructed by equation (4), (5) and (6) with an initial value solved by the previous step.

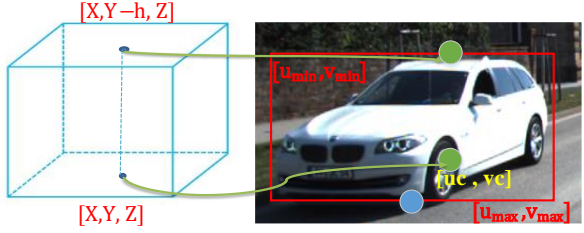

Fig. 6. Illustration of the relationship between the 3D bounding box's height and its projection on the image plane which are used to calculate the initial location of the 3D box.

## D. 2D detection correction based on 3D Estimation

In the autonomous driving scenarios, all the concerned objects, cars, cyclists and pedestrian are on the ground plane. And the RGB images in the KITTI dataset used in this paper are collected by the VW Passat station wagon which is equipped with four video cameras. All the cameras' heights above ground are 1.65 meters and the cameras' locations are used to establish the world coordinate system. In consideration of the characters about the autonomous driving scenarios, the image acquisition device, the acquisition environment and the 3D information of the objects obtained by the estimated 3D bounding box, we propose a 2D detections depurated method based on these recovered 3D physical information of the scenarios to filter out the false detection and false alarm in the 2D detections. We will describe in detail that the wrong detection in the 2D image may lead a 3D detection result having four aspects of unreasonable properties. The first two are about the 3D bounding box's attributes, and the latter two are about the relationship among the objects in view of the image. We use these characters to filter out the false alarm and false detection in the 2D detections.

First, the dimensions for the car category instances are low-variance to the average dimensions $[1.52, 1.64, 3.86]$. So the length, width and height of the estimated 3D bounding boxes should meet a certain range. If the 2D detection is wrong, this may lead to getting an abnormal car dimensions.

Second, for the objects in view of an image, their vertical locations are within certain range due to the image acquisition device is on the road surface of ground plane and with a fixed height of 1.65 meters. So the location of the car in the vertical direction is within a range. In [5], they set three kinds of height $[1.3888, 1.7231, 2.0574]$ for object proposals. We use these value as the initial centers of k-means cluster to the training dataset and get three cluster centers as $[1.28842871, 1.72911, 2.2184]$. So if the object's vertical location is far from this range, we will treat them as fallacious results. And the vertical locations of the adjacent objects change monotonously due to the light straight line dissemination characteristic and local plane and gradually changing characteristics of the roads.

Third, the characteristic of optical imaging system satisfy the characteristic that the projection size of the 3D bounding box is small in the long distance or depth and big on the contrary. Generally speaking, the projected heights of the 3D bounding boxes change monotonously with the distance based on the pinhole camera model. The same dimensions of the objects, their projected sizes will decrease with the increase of the distance which is also called depth, and can be denoted by the inequality as follows:

$$if\ Z1 > Z2\ and\ h1 = h2 + \varepsilon, \\ then\ v_{max1} - v_{min1} < v_{max2} - v_{min2} \quad (8)$$

where $Z1$ and $Z2$ represent the depths of two arbitrary objects, $\varepsilon$ denotes a small value within certain range. Specifically, the projected height of 3D boxes should change monotonously with the distance of the object from the image plane.

Finally, the adjacent objects may have similar depths. We use the IoU between two 2D detection boxes to determine whether they are adjacent or not. We consider the overlapped region in the left side and the right side of an object on the image plane as the two collected adjacent objects. If the depths of both the left adjacent 2D box and right adjacent 2D box of an object are far away from itself, we will treat this 2D box as a false detection.

If the detected 3D bounding box obviously violated one of these properties, especially the second property, we will directly remove the 2D detection and its corresponding 3D detection. For these are not easy to distinguished wrong detections, we use multi-properties to co-determination which one need to be removed from the detection results.

## IV. EXPERIMENTS

Experiments evaluated our framework are conducted on the real-world dataset, KITTI object detection benchmark [7] from driving street scenarios which including both 2D Object Detection Evaluation and 3D Object Detection Evaluation. It consists of 7481 training images and 7518 testing images in the dataset, and in each image, the object is annotated with observation angle (local orientation), 2D location, dimensions, 3D location, and global orientation. However, the annotated labels are only available in the training set, so our experiments are mainly conducted on the training set.

### A. Implementation Details

Our 3D properties estimation network was trained and tested on KITTI object detection dataset by the split used in [6]. We used the ImageNet pre-trained VGG-19 network [20] without its FC layers as backbone and added our 3D box properties branches behind it to complete each task. For the pre-training process, the cross-entropy loss was used for the classification and the smooth L1 loss is used for regression task, since it's less sensitive to outliers compared to the L2 loss. During training, each ground truth object was cropped and resized to 224x224 and fed into the 3D properties network. We filtered out those samples which are heavily truncated from the training set in case of the potential harm to the model and randomly apply mirroring and color distortions to the training images for data augmentation in order to make the network more robust to viewpoint varies and occlusions. Then the network was trained with SGD at learning rate of 0.0001 for 30 epochs with a batch size of 8 to get the final network parameters which were used for validation. We set the weighting factors of loss $w_{1:4} = [1, 4, 8, 4, 4]$. Fig. 5 showed some qualitative visualization of our result on KITTI validation set. For fair comparison, all our experiments were based on the MS-CNN [17] 2D detection method to produce 2D boxes and then estimate 3D boxes from 2D detection boxes whose scores exceed a threshold. For fair comparisons, we used the detection results reported by the authors. Since most works only released their result on cars, thus we made evaluation of our model on KITTI dataset focused on for car category like most previous works did. Our experiments were conducted on the setting of i7-6700 CPU, 16GB RAM, and NVIDIA GTX1080Ti GPU using Python and PyTorch [21] toolbox.

TABLE I
COMPARISON OF THE AVERAGE ORIENTATION SCORE (AOS, %), AVERAGE PRECISION (AP, %) AND ORIENTATION SCORE (OS) ON OFFICIAL KITTI DATASET FOR CARS.

| Method | Easy | | | Moderate | | | Hard | | |
|---|---|---|---|---|---|---|---|---|---|
| | AOS | AP | OS | AOS | AP | OS | AOS | AP | OS |
| Mono3D[5] | 91.01 | 92.33 | 0.9857 | 86.62 | 88.66 | 0.9769 | 76.84 | 78.96 | 0.9731 |
| 3DOP[13] | 91.44 | 93.04 | 0.9828 | 86.10 | 88.64 | 0.9713 | 76.52 | 79.10 | 0.9673 |
| SubCNN[7] | 90.67 | 90.81 | 0.9984 | 88.62 | 89.04 | 0.9952 | 78.68 | 79.27 | 0.9925 |
| Deep3DBox [8] | 97.50 | 97.75 | 0.9974 | 96.30 | 96.80 | 0.9948 | 80.40 | 81.06 | 0.9919 |
| Ours | **97.57** | 97.75 | **0.9981** | **96.50** | 96.80 | **0.9970** | **80.45** | 81.06 | **0.9925** |

TABLE II
COMPARISON WITH THE STATE-OF-THE-ART METHODS USING THE METRIC OF 3D AVERAGE PRECISION(%) ON KITTI DATASET TO EVALUATE THE 3D DETECTION ACCURACY FOR CAR CATEGORY. THE BEST RESULT OF EACH COLUMN IS HIGHLIGHTED WITH BOLD FONT.

| Method | Type | Time | $AP_{3D}$ - IoU=0.5 | | | $AP_{3D}$ - IoU=0.7 | | |
|---|---|---|---|---|---|---|---|---|
| | | | Easy | Moderate | Hard | Easy | Moderate | Hard |
| 3DOP[13] | stereo | 3s | 46.04 | 34.63 | 30.09 | 6.55 | 5.07 | 4.10 |
| Mono3D[5] | mono | 4.2s | 25.19 | 18.20 | 15.52 | 2.53 | 2.31 | 2.31 |
| Deep3DBox [8] | mono | - | 27.04 | 20.55 | 15.88 | 5.85 | 4.10 | 3.84 |
| FQNet [18] | mono | 3.3s | 28.98 | 20.71 | 18.59 | 5.45 | 5.11 | 4.45 |
| GS3D [19] | mono | 2.3s | 30.60 | 26.40 | 22.89 | 11.63 | 10.51 | 10.51 |
| Ours | mono | 0.258s | **31.45** | 24.52 | **20.48** | **13.79** | **11.38** | **10.95** |

TABLE III
$AP_{3D}$ FOR THE PEDESTRIAN AND CYCLIST CATEGORY ON KITTI VALIDATION DATASET

| Method | Easy | | | Moderate | | | Hard | | |
|---|---|---|---|---|---|---|---|---|---|
| | AOS | AP | OS | AOS | AP | OS | AOS | AP | OS |
| 3DOP[5] | 70.13 | 78.39 | 0.8946 | 58.68 | 68.94 | 0.8511 | 52.32 | 61.37 | 0.8523 |
| Mono3D[4] | 65.56 | 76.04 | 0.8621 | 54.97 | 66.36 | 0.8283 | 48.77 | 58.87 | 0.8284 |
| SubCNN[7] | 72.00 | 79.48 | 0.9058 | 63.65 | 71.06 | 0.8957 | 56.32 | 62.68 | 0.8985 |
| Deep3DBox [8] | 69.16 | 83.94 | 0.8239 | 59.87 | 74.16 | 0.8037 | 52.50 | 64.84 | 0.8096 |
| Ours | 75.33 | 83.94 | 0.8974 | 66.08 | 74.16 | 0.8910 | 57.42 | 64.84 | 0.8856 |

TABLE V
ABLATION STUDY OF 3D DETECTION RESULTS FOR CAR CATEGORY ON KITTI VALIDATION DATASET.

| method | $AP_{3D}$ Iou =0.5 | | | $AP_{3D}$ Iou =0.7 | | |
|---|---|---|---|---|---|---|
| | easy | moderate | hard | easy | moderate | hard |
| cbf | 31.47 | 23.11 | 19.17 | 7.95 | 5.66 | 5.31 |
| vp | 30.24 | 19.85 | 17.11 | 8.19 | 5.50 | 5.10 |
| cbf /vp | 33.03 | 25.68 | 21.48 | 9.69 | 7.98 | 6.56 |
| focal loss cbf /vp | 31.45 | 24.52 | 20.48 | 13.79 | 11.38 | 10.95 |

*B. 3D Bounding Box Evaluation*

We compared our proposed method with 6 recently proposed state-of-the-art monocular based 3D object detection methods on the KITTI benchmark, including 3DOP [13], Mono3D [5], SubCNN [7], Deep3DBox [8], FQNet [18] and GS3D [19] for KITTI cars. These results are evaluated based on three levels of difficulty: Easy, Moderate, and Hard, which is defined according to the minimum bounding box height, occlusion degree and truncation grade. We evaluated the orientation, dimensions and the overall performance of the 3D bounding boxes.

**Training data requirements.** As all 3D properties we learned were appearance related, thus we could overcome the downside of Deep3DBox[8] that needed to learn the parameters for the fully connected layers and required more training data than methods that using additional information, such as segmentation, depth. By adding two appearance-related tasks, we got competitive performance with [8] by less training data.

**KITTI orientation accuracy.**

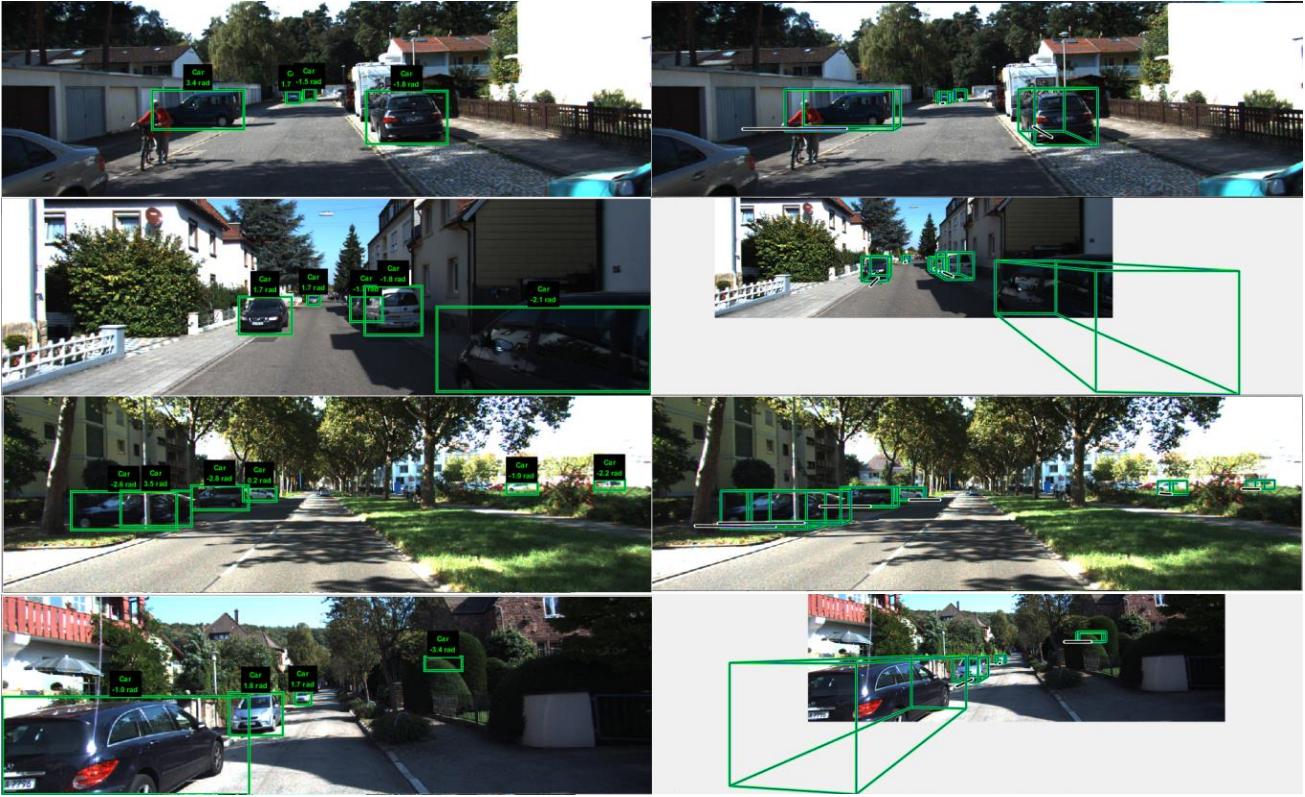

Fig. 7. Result visualization of the 2D detection boxes (left, ported from [8]) and estimated 3D box projections (right) for cars on KITTI validation dataset by our cascaded geometric constraints without using the 2D detections filtration stage. The black lines attached to each 3D box represent the orientation of objects (start from the center of bottom face to the front of objects).

TABLE IV
COMPARISONS OF THE AVERAGE ERROR OF DIMENSION ESTIMATION WITH STATE-OF-THE-ART METHODS ON THE KITTI VALIDATION DATASET. THE NUMBER IS THE SMALLER THE BETTER.

| Method | Dims errors |
|---|---|
| Mono3D[4] | 0.4251 |
| 3DOP[13] | 0.3527 |
| Deep3DBox[8] | 0.1934 |
| FQNet[18] | 0.1698 |
| Our Method | 0.1663 |

Average Orientation Similarity (AOS) is the official 3D orientation metric of the KITTI dataset which is described in [1] and multiplies the average precision (AP) of the 2D detector with the average cosine distance similarity for azimuth orientation is calculated to evaluate the performance of orientation estimation. The ratio of AOS over Average Precision (AP) called OS is defined in [8] which is representative of how well each method performs only on orientation estimation, while factoring out the 2D localization performance. The AOS is first published in [8] as assessment criteria and our method is first among all non-anonymous methods for car examples on the KITTI dataset. As shown in Table 1, our method using exactly the same 2D detector outperformed the baseline and other monocular images based 3D detection methods on orientation estimation for cars. Our method even outperformed Deep3DBox [8] for cyclist categories despite having similar 2D AP as shown in Table 3. On the KITTI detection dataset, 2 bins was achieved better performance than 8 bins in our work as it decreased the training data amount for each bin. We also conducted experiments with different numbers of neurons in the fully connection layers (see Table 6) and found that increasing the number of neurons in the FC layers further yielded some limited gains even beyond 256.

**KITTI 3D bounding box metric.**

The orientation estimation precision evaluated only part of 3D bounding box's parameters. To evaluate the accuracy of the rest, we introduced 3 metrics, on which we compare our method against FQNet [18] for KITTI cars. The first metric was the average error in estimating the dimensions of the 3D objects which was defined as: $E_d = 1/N \sum_{i=1}^{N} \sqrt{\Delta w_i^2 + \Delta h_i^2 + \Delta l_i^2}$. We found the corresponding object in the ground truth which were closest to the detection result for computing $E_d$. We only compared our method with Mono3D [4], 3DOP [13], Deep3DBox [8] and FQNet [18] which have provided their experimental results. Our results were summarized in Table 4. We could see that our methods had the lowest estimation error with an average dimension estimation error of about 0.1663 meters, which demonstrated the effectiveness of our collaborated appearance-related properties regression modules.

We further evaluated the 3D Object Detection performance with the 3D AP metric. 3D AP is the KITTI official metric which is used to evaluate the overall advantages of 3D bounding box estimation. The threshold of 3D Intersection over Union (IoU) with 0.7 and 0.5 are both used to determine a detection result is successful or not. As indicated in Table 2, our method performed relatively well compared to these most related monocular image based methods[8, 18] and even ranked first among purely monocular based methods with the IoU threshold of 0.7. From Table 2, we can see that our method outperformed Mono3D [8] and Deep3DBox [32] by a significant margin of about 7% improvement and even outperformed stereo-based 3DOP when 3D IoU threshold is set to 0.7. Since 3DOP [9] is a stereo-based method that can obtain depth information directly, so its performance is much better than pure monocular based methods with the IoU threshold of 0.5. The cost time of inference was also shown in this table, which demonstrated the efficiency of our method. Our method did not rely on computing additional features such as stereo, semantic or instance segmentation, depth estimation as well and did not need complex postprocessing as in [24] and [23]. The ablation study of the contribution of vp, cbf and cascaded constraint vp/cbf were shown in Table.5. When we used the focal loss for category classification, we would get even better performance.

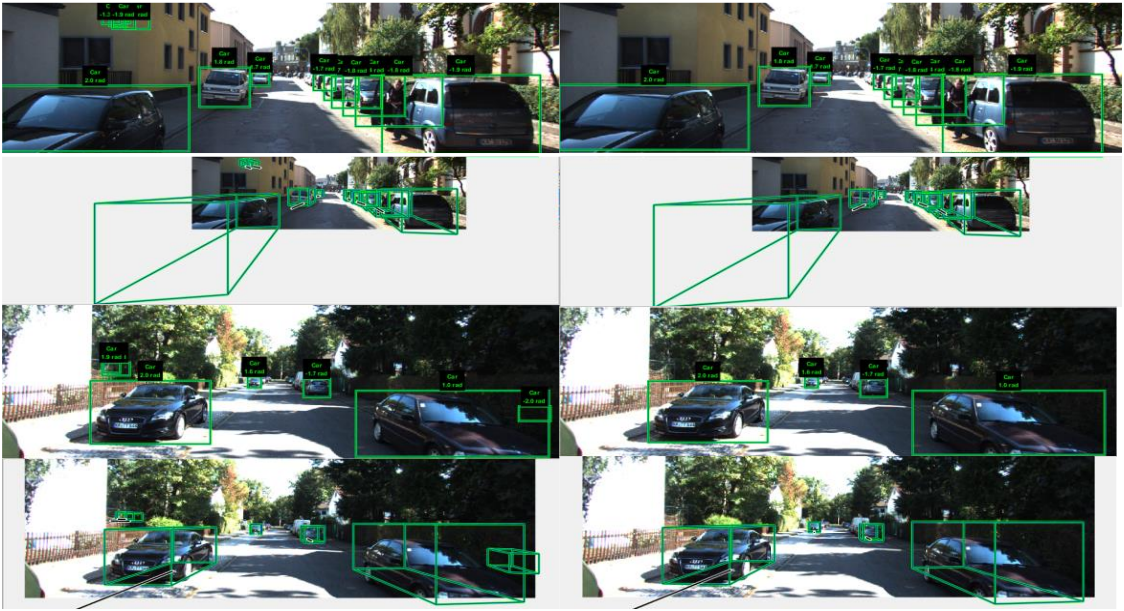

Fig. 8. Examples of these obviously abnormal detections which can be easily removed by one of the physical world prior information.

TABLE VI
THE EFFECT OF NUMBER OF NEURONS IN FC LAYERS FOR ORIENTATION ESTIMATION

| FC | 64 | 128 | 256 | 512 | 1024 |
|---|---|---|---|---|---|
| OS | 0.98472 | 0.98553 | 0.99946 | 0.9996 | 0.9995 |

As image-based 3D detection methods had many drawbacks in spatial localization, 3D AP score was relatively lower than that 2D detectors obtain on the corresponding 2D metric. This was due to the fact that 3D estimation was a more challenging task, especially as the distance to the object increases. For example, if the car was 50m away from the camera, a translation error of 2m corresponds to about half the car length would hard to identify on the image plane. Our method handles increasing distance well, as shown in Fig. 7. The evaluation shown that regressing the CBF and viewpoints made a difference in all the 3D metrics. All the quantitative results given in our paper were only obtained by the cascaded geometric constraints without filtering procedure as the filtering procedure could be clearly observed in the image.

**Qualitative Results:** We had drawn the projected 3D detection boxes on 2D image plane for better visualization. Fig. 7 was the examples of qualitative detection results by cascaded geometric constraints method without filtering wrong 2D detections on the scenes of KITTI dataset. The results in Fig. 7 showed that our approach could deal with the truncated object detect the 3D object well and achieve high-precision 3D perception in autonomous driving scenarios with only one monocular image as input. If the 2D bounding box was closed to the image boundary and less than 10 pixels, we treated is as truncated object.

### C. Filtering 2D Detection based on 3D Box Estimation

By the high precision 3D detector of our geometric cascade constraints, we can filter out the wrong 2D detections with a high reliable. There are 1% of 2D detections are obviously wrong detections and easily identified by the location of the vertical direction whether normal or not for cars category in the autonomous driving scenarios. In Fig. 8 was shown some examples of easily identified false alarms.

As the projected size decreases approximately linearly with distance and the nearby objects may have similar depth, these could be used to filtering out almost 5% of more complicated wrong detections. From Fig.9 we could see that by using the 3D result combined with the physical information, more complex false alarm and false detection also could been removed. Once the wrong 2D detections are eliminated, its correspondence 3D detections will also be abandoned.

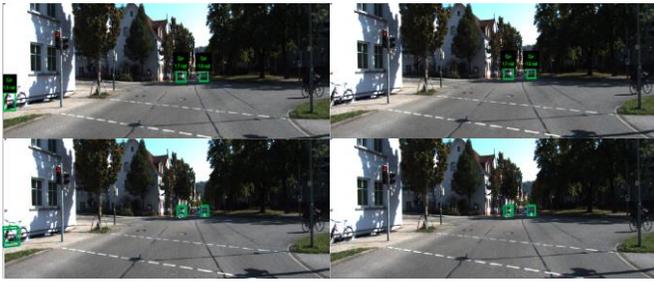

Fig. 9. Examples of these more complex abnormal detections which can be removed by several physical prior information. The left two images contain a false detection in the lower left which are hard to distinguish. We jointly use the depth, location and dimensions to find the false detection.

## V. CONCLUSION AND FUTURE WORK

In this paper, we have proposed a novel method using deep CNN to regress another two appearance-related 3D properties, viewpoints classification and the center projection of the 3D box's bottom face, and using these two properties to construct a cascaded geometric constraints model which is used to solve a more precision 3D location. Then we use the recovered 3D physical world information to further depurate the 2D detections.

Experiments demonstrated that our cascaded geometric constraints method is not only less time and computational resources consuming than the baseline algorithm which makes this method with higher applicability, but also can deal well with the truncated objects been by the image plane. By the post-processing steps, we can wipe out most false alarm and false detection in 2D and 3D detections. Although our method have achieved better performance, it remains a problem that heavily depend on 2D detection performance which we hope to solve it and make our method less sensitive to the 2D detection. And we also expect to extend our monocular 3D object detection method for monocular 3D object tracking in the future.